\documentclass[runningheads]{llncs}

\usepackage{eccv}

\usepackage{eccvabbrv}

\usepackage{afterpage}

\usepackage{graphicx}
\usepackage{booktabs}
\usepackage{cuted}
\usepackage{amssymb}
\usepackage{pifont}
\newcommand{\xmark}{\ding{55}}
\usepackage{multirow, makecell}
\newcommand{\thinmidrule}

\usepackage[accsupp]{axessibility}  

\usepackage[pagebackref,breaklinks,colorlinks,citecolor=eccvblue]{hyperref}
\usepackage{hyperref}

\usepackage{orcidlink}
\usepackage{multicol}

\begin{document}

\newcommand{\mycomment}[1]{}

\title{LingoQA: Visual Question Answering for Autonomous Driving} 


\author{Ana-Maria Marcu* \orcidlink{0009-0009-6494-6018} \and
Long Chen*\orcidlink{0000-0002-4985-9516} \and
Jan Hünermann* \and Alice Karnsund* \and Benoit Hanotte* \and Prajwal Chidananda \and Saurabh Nair \and Vijay Badrinarayanan \and Alex Kendall \and Jamie Shotton \and Elahe Arani \and Oleg Sinavski* }

\authorrunning{A-M. Marcu, L. Chen, J. Hünermann, A. Karnsund, B. Hanotte et al.}

\institute{Wayve Technologies, 230-238 York Way, London N7 9AG, United Kingdom
\email{research@wayve.ai} \\
\url{https://wayve.ai/science/} \\
}

\maketitle

\begin{center}
\small\textsuperscript{*}Equal contributions
\end{center}

\begin{abstract}
   We introduce LingoQA, a novel dataset and benchmark for visual question answering in autonomous driving. The dataset contains 28K unique short video scenarios, and 419K annotations. Evaluating state-of-the-art vision-language models on our benchmark shows that their performance is below human capabilities, with GPT-4V responding truthfully to 59.6\% of the questions compared to 96.6\% for humans. For evaluation, we propose a truthfulness classifier, called Lingo-Judge, that achieves a 0.95 Spearman correlation coefficient to human evaluations, surpassing existing techniques like METEOR, BLEU, CIDEr, and GPT-4. We establish a baseline vision-language model and run extensive ablation studies to understand its performance. We release our dataset and benchmark\footnote{\href{https://github.com/wayveai/LingoQA}{https://github.com/wayveai/LingoQA}} as an evaluation platform for vision-language models in autonomous driving.
   \\\keywords{Autonomous Driving \and Question Answering \and Vision-language}
\end{abstract}

\begin{figure*}[h]
    \centering
    \includegraphics[width=1.0\textwidth]{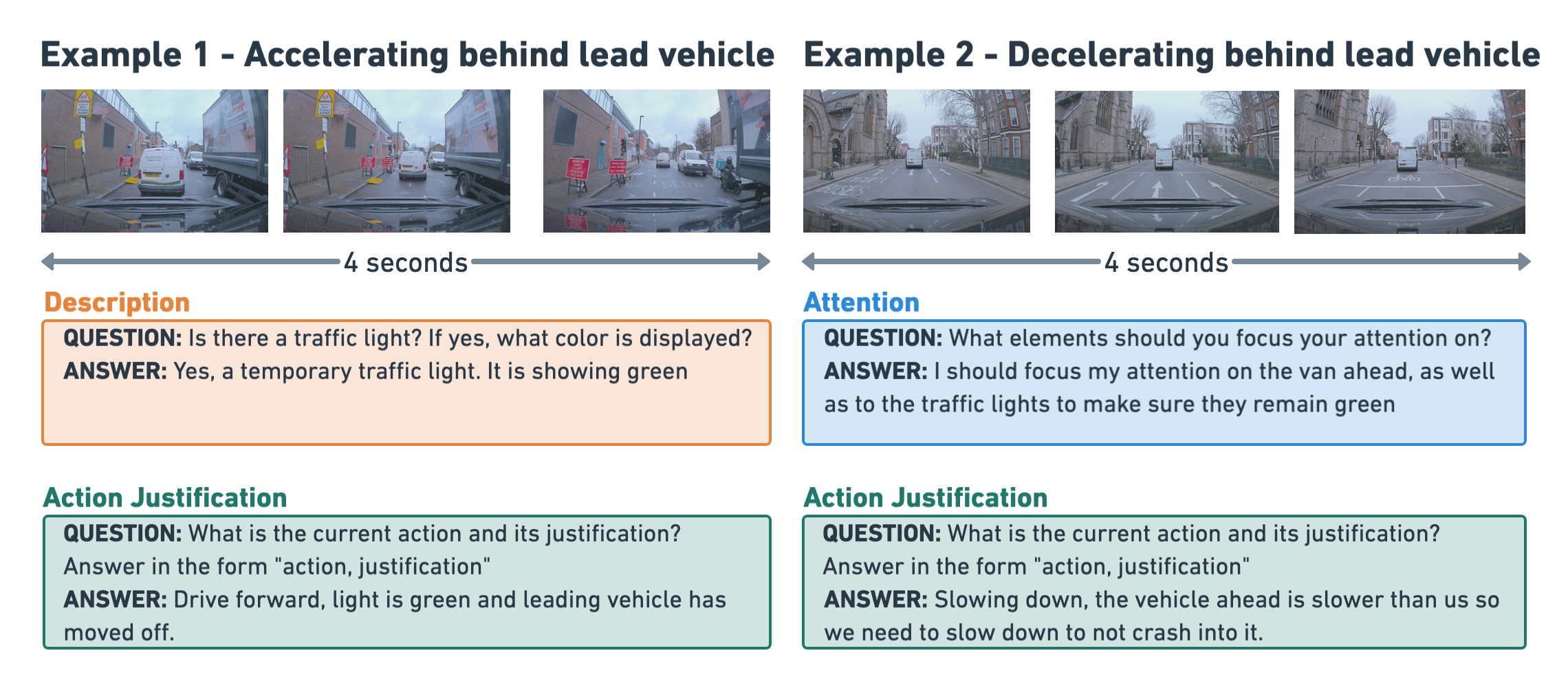}
    \captionof{figure}{Examples from the LingoQA benchmark designed to evaluate whether models can describe the elements in the scene (description), predict what future action should be taken (action) and why (justification) and explain what they are paying attention to (attention).}
    \centering
    \label{fig:lingo_examples}
\end{figure*}

\section{Introduction}
\label{sec:intro}

Communication plays a pivotal role in naturally fostering trust among individuals. However, establishing trust between users and agents remains a significant challenge within the field of artificial intelligence. Recent studies have indicated that articulating explicit reasoning steps can significantly enhance user confidence \cite{PAVEpoll}, in addition to improving the capabilities of Large Language Models (LLMs) \cite{wei2023chainofthought}. The need for textual justifications remains critical, particularly in safety-critical domains where technology adoption hinges upon this factor \cite{kim2018textual}.

Consider the domain of end-to-end autonomous driving \cite{chen2023endtoend}, where the driving policy is often executed through deep neural networks processing camera inputs to generate control commands. Recent strides in VLMs have solidified transformers as multimodal learners, showcasing remarkable performance in tasks such as Visual Question Answering (VQA) and underscoring their proficiency in acquiring robust representations for complex tasks \cite{chib2023recent}. Integrating Vision-Language Models (VLMs) into the field of autonomous driving holds the promise of enhancing user trust in these systems.

Our focus is on vision-only end-to-end autonomous driving, aiming to bridge the gap between data-driven decision-making and user trust. We introduce LingoQA, a benchmark designed for autonomous driving VQA, utilizing a novel dataset comprising more than 419k QA pairs. Distinguished by its free-form approach to questions and answers, this dataset broadens the scope of autonomous driving QA, encompassing reasoning and action justifications. Additionally, we publish a comprehensive evaluation suite consisting of 1,000 examples. At the core of our benchmark lies a novel evaluation metric based on a learned text classifier called \textit{Lingo-Judge}, inspired by GPT-Judge used in TruthfulQA \cite{lin2022truthfulqa}.  We perform rigorous studies correlating automatic metrics to human preferences and find that Lingo-Judge achieves a 0.950 Spearman and 0.993 Pearson correlation coefficient, surpassing existing automated labelling techniques like METEOR \cite{banerjee-lavie-2005-meteor}, BLEU \cite{papineni-etal-2002-bleu}, CIDEr \cite{vedantam2015cider}, and GPT-4 \cite{openai2023gpt4} on our benchmark, while being fast enough for frequent runs during training and development. The evaluation code and the weights for the classifier will be released with the paper to support robust benchmarking of visual question answering in autonomous driving. 

Equipped with this evaluation toolkit, we conducted a comprehensive empirical study on key components and their ablations in VLMs for autonomous driving. Our findings in Section \ref{sec:empirical_evaluation} indicate that the most effective approach involves partially fine-tuning the attention layers of our vision-language model equipped with Vicuna-1.5-7B \cite{vicuna2023}, on both Action and Scenery datasets. This process involves using 5 video frames over 4 seconds and a late video fusion technique. Our collective work, spanning the LingoQA benchmark, the visual instruction-tuning dataset, and the innovative evaluation metric, aims to propel the domain of language-augmented autonomous driving, laying a robust foundation for subsequent research and development endeavors. To summarise the main contributions of this paper:

\begin{itemize}
\item \textbf{LingoQA Benchmark}: We introduce a novel benchmark for autonomous driving VQA using a learned text classifier for evaluation. It outperforms existing metrics, including GPT-4, with a Spearman coefficient of 0.950 indicating a strong correlation with human evaluation.

\item \textbf{LingoQA Dataset}: Our 419.9k QA pair dataset stands out with its free-form questions and answers, covering not just perception but also driving reasoning from the drivers directly, broadening the scope of autonomous driving datasets.

\item \textbf{LingoQA Baseline}: Through testing of various vision-language components on LingoQA, we find that the most effective approach involves partially fine-tuning the attention layers of our vision-language model equipped with Vicuna-1.5-7B \cite{vicuna2023} and a late video fusion technique. We establish a new baseline for this field with an identified model combination. Example outputs from the model are shown in Figure \ref{fig:lingo_examples}.
\end{itemize}

\section{Related Work}

\subsection{Language in Autonomous Driving}
Modern autonomous vehicle software relies heavily on artificial intelligence models \cite{bansal2018chauffeurnet, gao2020vectornet, hawke2021reimagining, NEURIPS2022_827cb489}. This, together with the increased number of such vehicles on the road, poses a fundamental challenge in terms of interpretability in the decision-making process \cite{BARREDOARRIETA202082}. Understanding why a decision is made is crucial for understanding areas of uncertainty, building trust, enabling effective human-AI collaboration, and ensuring safety \cite{HumanInteractions}. In a survey conducted by Partners for Automated Vehicle Education (PAVE) in 2020 \cite{PAVEpoll}, 60\% of participants stated that they would trust AVs more if they better understood the underlying process of the system. To establish trust with the general public, the systems must be explained in a human-interpretable way, such as through language and visual explanations \cite{BARREDOARRIETA202082}. 

The field of autonomous driving has been embracing the opportunity to make driving models more trustworthy for their users using visual attention methods \cite{jain2019attention} or textual explanations \cite{kim2018textual}. The early explorations of GPT3.5 \cite{sha2023languagempc,gptdriver} and GPT4-V \cite{wen2023road} on autonomous driving scenarios show that LLMs/VLMs demonstrate superior performance in scene understanding and causal reasoning compared to existing autonomous systems. Works such as ADAPT \cite{jin2023adapt} and LLM-Driver \cite{chen2023driving} propose multi-task learning frameworks for jointly predicting language and control outputs. Inspired by progress in large language models \cite{touvron2023llama,vicuna2023,zhang2022opt,openai2023gpt4}, vision-language models \cite{beit3,vlmo,simvlm,yang2022unified,Yu2022CoCaCC,zhang2022glipv2,pali2023,flamingo2022,blip2,openai2023gpt4,liu2023visual,dai2023instructblip} and multi-modal transformers for robotics \cite{driess2023palme,brohan2023rt1,brohan2023rt2} our work incorporates language to autonomous driving. Closely related to our proposed baseline is DriveGPT \cite{xu2023drivegpt4}, proposing a multi-modal vision-language-action model that tokenizes videos, as well as text and control actions. 
 
\subsection{Evaluation Metrics}
Progress has been relatively slow in developing vision-language models for autonomous driving, with only a few works aiming to quantitatively improve upon prior work \cite{kim2017interpretable,jin2023adapt,xu2023drivegpt4}. A key challenge consists of automated, reproducible evaluation metrics that are highly correlated with human ratings, particularly due to the inherent complexities in assessing natural language. ADAPT \cite{jin2023adapt} reports human feedback in addition to standard natural language generation metrics, while DriveGPT \cite{xu2023drivegpt4} reports ChatGPT ratings. Automated methods such as BLEU \cite{papineni-etal-2002-bleu}, METEOR \cite{banerjee-lavie-2005-meteor}, ROUGE \cite{lin-2004-rouge} show weak alignment with human feedback \cite{vedantam2015cider}. CIDEr \cite{vedantam2015cider} is also based on \textit{n-gram} level similarity, as opposed to capturing the correctness of an answer based on its meaning. Newer evaluation metrics using ChatGPT have shown improvement in the area of sentence understanding, while still having limitations, such as providing high scores to elaborate, eloquent, but incorrect sentences \cite{llmleaderboard}. Evaluation based on human feedback is subjective and difficult to reproduce. In this work, we address this challenge by introducing a novel VQA benchmark for autonomous driving that checks for factual correctness and is highly correlated to human correctness ratings on our proposed evaluation dataset. 

\subsection{Datasets for Autonomous Driving}
Recent advances in generative AI have been underpinned by training with increasingly large and diverse internet-scale datasets. \cite{flamingo2022} \cite{blip2} This has brought into light the need for evaluation benchmarks and datasets that focus not only on specific tasks, but on reasoning areas, such as descriptive and predictive reasoning. \cite{patraucean2023perception} Prior works, such as the CausalVQA benchmark \cite{li2022from} and the Perception Test \cite{patraucean2023perception}, a comprehensive benchmark for vision-language foundation models, probe the validity of the model representations through question answering. Autonomous driving datasets have been focused on commentary \cite{kim2018textual,xu2020explainable} or constructed around existing object detection datasets \cite{qian2023nuscenes, deruyttere2019talk2car}. Datasets such as NuScenesQA \cite{qian2023nuscenes}, contain simple language outputs of on average one word per question that do not tackle the more challenging reasoning problem.

Our proposed dataset LingoQA addresses the existing gap in autonomous driving as it contains a diverse set of questions related to \textit{driving behaviours} and \textit{scenery} in addition to perception questions related to object presence and positioning. The evaluation dataset probes areas such as description, counting, localisation, anticipation, attention, and action justification. This dataset has the strength of being diverse with respect to the language used while being grounded in human reasoning. Examples of the complex questions and answers existent in the dataset are provided in Appendix \ref{appendix:dataset_examples}.

\section{LingoQA Benchmark}

In this section, we introduce LingoQA, a benchmark to evaluate autonomous driving reasoning through video question-answering. The benchmark consists of an automated evaluation metric and the corresponding datasets for evaluation and fine-tuning.

\subsection{Evaluation Metric}\label{sec:evaluation_metric}
 
Evaluating open-ended textual dialogues is a challenging task. Quite often the correct answers are ambiguous, subjective, or even not attainable. The most common language-based metrics for evaluating question-answering models in autonomous driving \cite{xu2023drivegpt4, jin2023adapt, kim2018textual} are BLEU \cite{papineni-etal-2002-bleu}, METEOR \cite{banerjee-lavie-2005-meteor} and CIDEr \cite{vedantam2015cider}, despite their known limitations, such as relying heavily on the \textit{n-gram} frequency as opposed to the underlying meaning of the answer. To address these limitations, we set ourselves the challenge to develop an \textit{automated}, \textit{non-visual} evaluation method for free-form language answers from vision-language models which checks correctness independent of phrasing against a ground truth answer and which is \textit{highly correlated} with human ratings. 

\paragraph{N-Gram Matching Metrics.} Common language-based metrics such as BLEU \cite{papineni-etal-2002-bleu}, METEOR \cite{banerjee-lavie-2005-meteor} and CIDEr \cite{vedantam2015cider} are still the most common metrics used for evaluating captioning in autonomous driving, despite their known limitations, such as relying heavily on the \textit{n-gram} frequency as opposed to the underlying meaning of the answer. The BLEU \cite{papineni-etal-2002-bleu} metric computes the \textit{n-gram} based precision of the candidate sequence compared to the reference sequence. Similarly, ROUGE \cite{lin-2004-rouge} computes the recall-based n-gram, while METEOR \cite{banerjee-lavie-2005-meteor} computes the F-measure. CIDEr \cite{vedantam2015cider} is based on using TF-IDF to provide a lower weight to terms that are commonly reported in the corpus. 

\begin{table}[tb]
\begin{center}
\caption{\textbf{Lingo-Judge performance.} Correlation with human ratings, validation accuracy, and time taken to run of our proposed LingoQA evaluation metric compared to previous language-based metrics. All metrics use textual ground truth and have no access to vision information. Further examples are presented in Appendix \ref{classifier:examples}.}  
\resizebox{0.7\linewidth}{!}{
\begin{tabular}{|l|c|c|c|c|}
\hline
& Pearson & Spearman & Val Acc. [\%] & Time [sec] \\
\hline
$\text{Lingo-Judge}$ & \textbf{0.993} & \textbf{0.950} & \textbf{95.0} & 10.5 \\
$\text{GPT-4 with CoT}$  & 0.990 & 0.932 & 91.2 & 3016.0 \\
$\text{GPT-4}$ \cite{openai2023gpt4} & 0.988 & 0.941 & 90.6 & 812.4 \\
$\text{BLEU}$ \cite{papineni-etal-2002-bleu} & 0.881 & 0.835 & - & \textbf{0.1} \\
$\text{METEOR}$ \cite{banerjee-lavie-2005-meteor} & 0.891 & 0.876 & - & 8.0\\
$\text{CIDEr}$ \cite{vedantam2015cider} & 0.878 & 0.853 & - & 0.2 \\
\hline
\end{tabular}}
\label{tab:correlation}
\end{center}
\end{table}

\paragraph{GPT-4 based evaluation.} Inspired by the G-Eval metric \cite{liu2023geval}, we used GPT-4 to evaluate answers on a larger scale. Given a question and answer pair from the test set and a model's answer, we ask GPT-4 to evaluate whether the model's answer corresponds to a human's answer. Notice that it does not make use of any visual input.
We experimented with prompts and methods achieving good quality of judgements. We achieved the highest accuracy by employing chain-of-thought prompting where we ask GPT-4 to first come up with an evaluation strategy before grading a model’s answer. However, as shown in Table \ref{tab:correlation}, this leads to increased inference time. Further details are provided in Appendix \ref{appendix:chatgpt}. 
Unfortunately, we found GPT-4 based evaluation impractical to use as a main development and training metric due to the time required to evaluate answers on our relatively small evaluation dataset (from 13min up to 50min for a single evaluation due to the API rate limit).

\paragraph{Lingo-Judge.}  Given these limitations and inspired by TruthfulQA GPT-Judge \cite{lin2022truthfulqa}, we pursued an alternative approach using a learned text classifier, dubbed Lingo-Judge, which estimates the correctness of model answers. We measure the correctness of model predictions as an accuracy using a small transformer-based text classifier that takes in a question, the human's, and the model's answer and outputs a probability that the model's answer is correct. Please note, Lingo-Judge does not receive video input and must rely only on the supporting human's answers. For every question, we run Lingo-Judge on all combinations of \textit{(ground-truth answer, predicted answer)} and take the maximum correctness estimate, as shown in Equation \ref{eq:judge}, where $S$ is the score per sample. We found that this recipe yields the best predictive power provided enough diversity of human answers in our evaluation dataset.

\begin{equation}
\label{eq:judge}
    S = \max_{j \in \{0, 1\}}F_{\mathrm{Judge}}(\mathrm{prediction},  \mathrm{ground\_truth}[j])
\end{equation}

The architecture of the classifier is a DeBERTa-V3 \cite{he2023debertav3} language model, fine-tuned with LoRA \cite{hu2021lora}. The classification score is predicted using a linear head on top of the class token output. We fine-tuned the model on a diverse dataset of model predictions from early experiments, where questions and ground truth answers come from our evaluation dataset and the correctness target is labeled by human annotators. On top of this initial dataset, we iteratively improved the classifier using active learning by correcting the wrong predictions of discarded models and adding corrections to the training dataset. On a held-out test set, we find that the binary classification accuracy of the classifier is 95\%.

In comparison to metrics such as CIDEr, which provide a \textit{system-level} performance metric, the classifier provides a probability of correctness for each of the model predictions, meaning that it provides metrics at the \textit{sample} level. Examples are provided in the Appendix \ref{classifier:examples}. This means that 100\% classifier accuracy is easy to interpret. The classifier allows us to compute metrics during training, running over our full evaluation dataset in 10 seconds using an A100 GPU.

\paragraph{Correlation to human ratings.}

We studied empirical \textit{correlation} of various metrics with human judgments. Several human annotators assigned a scalar score [0, 1] to the inference outputs of 17 different models which can be interpreted as the probability that the response correctly addresses the question \cite{lin2022truthfulqa}. Notably, this process takes several days, highlighting the need for an automated evaluation metric that provides faster development feedback. The final human score of each model is the average of all inference output scores. Further details regarding the methodology for the correlation analysis are in the Appendix \ref{correlation:study}.

The \textit{Spearman rank correlation} coefficient of our automated metric, Lingo-Judge, with human scores is 0.95, and the \textit{Pearson correlation} coefficient is 0.993. These values are considerably higher compared to other natural language evaluation metrics and GPT-4, as detailed in Table \ref{tab:correlation}. Our analysis demonstrates that Lingo-Judge accurately mirrors human judgments, outperforming existing metrics such as BLEU, METEOR, and CIDEr, as well as GPT-4 with and without chain-of-thought prompting. This indicates that Lingo-Judge can effectively serve as a proxy for human labelling, which is particularly significant given the stagnant nature of metrics in autonomous driving since the introduction of the CIDEr metric in 2015. Notably, despite their limitations, prominent models like ADAPT \cite{jin2023adapt} and DriveGPT \cite{xu2023drivegpt4} still use BLEU, METEOR, and CIDEr metrics and report ChatGPT ratings without analyzing their correlation to human preferences. Our work fills this gap by providing a reliable benchmark that better reflects human preferences.

\subsection{Datasets} \label{subsec:dataset}
We created a collection of datasets for bringing language to autonomous driving. The total dataset size is 419.9k question-answer pairs, where a single data sample consists of a 4-second video clip at 1Hz. The total size of the dataset is about 10x larger than BDD-X \cite{kim2018textual}, as shown in Table \ref{tab:dataset_comparisons}. Compared to prior datasets such as NuScenesQA \cite{qian2023nuscenes}, our dataset contains reasoning pairs in addition to object presence, description, and localisation. The answers are also free-form and more complex, with an average answer length of 17.2 words versus 1.0 words in NuScenesQA. Examples of question answers pairs from LingoQA are shown in Appendix \ref{appendix:dataset_examples}. 

\begin{table}
    \caption{\textbf{Dataset comparison.} The dataset that we introduce has a similar size to other driving-related while having a much higher diversity. A scenario refers to a unique short video sequence, that may have multiple annotations attached to it.}
    \resizebox{0.85\linewidth}{!}{
    \begin{tabular}{|l|c|c|c|c|c|}
    \hline
    & \textbf{Scenarios} & \textbf{Annotations} & \textbf{QA} &  \textbf{Captioning} & \textbf{Video length [sec]} \\
    \hline
    Rank2Tell \cite{sachdeva2023rank2tell} & 118 & $>118$ & \xmark & \checkmark & 20 \\
    BDD-OIA \cite{xu2020explainable} & 22.9k & 35k & \xmark & \checkmark & 5  \\
    BDD-X \cite{kim2018textual} & 6.9k & 26k & \xmark & \checkmark & 40 \\
    NuScenesQA \cite{qian2023nuscenes} & 34k & 460k & \checkmark & \xmark & 20 \\
    DriveLM \cite{drivelm2023} & 30k & 443k & \checkmark & \checkmark & 20 \\
    \textbf{LingoQA} & 28k & 419.9k & \checkmark & \checkmark & 4 \\
    \hline
  \end{tabular}}
  \centering
  \label{tab:dataset_comparisons}
\end{table}

Our labeled autonomous driving training dataset consists of two complementary parts: the \textit{action} dataset and the \textit{scenery} dataset. The breakdown of the datasets is shown in Table \ref{tab:action_scenery}.

\paragraph{Action dataset.} The action dataset was created from recorded driving sessions featuring notable events where the car's behavior changes, such as decelerations, accelerations, lane changes, narrow gaps, and turns. These events were labeled by driving operators with concise, high-level descriptions of the situations and behaviors (e.g., "following lane, pedestrian on a zebra crossing, should stop"). Additional metadata was included from various perception systems, covering traffic light presence, vehicle and pedestrian detection, weather conditions, and other data like speed, steering wheel position, and road type. Using this information, we developed prompt templates for (1) describing the current action and its justification and (2) generating example questions and hints for expected answers. We then used these prompts with GPT-3.5 to rephrase, answer, and extend the example questions based on the provided action descriptions and answer hints. Events were rebalanced by categorizing them based on actions and behavioral policies, and up to 500 events were sampled from each category, resulting in 24,577 video snippets with 167,774 question/answer pairs.

\paragraph{Scenery dataset.} The scenery dataset was designed to complement the action dataset by focusing on fine-grained perception-related questions. It was created by densely and thoroughly annotating three 30-minute driving sessions using the ELAN video annotation software. \cite{elan} During these sessions, brief captions were provided in approximately 15 different categories, including driver actions, justifications, attention, observations of vehicles, pedestrians, road elements (such as traffic lights, traffic islands and intersections), and environmental details (like weather and buildings). Annotations were collected every second (1 fps) for every frame to build textual descriptions that included driver actions, justifications, and observations. Unlike the Action dataset, where GPT-3.5 was used only for question rephrasing, GPT-4 was employed for the Scenery dataset to generate questions and answers. GPT-4 was prompted with chain of thought specifically targeting perception questions, resulting in a diverse and high-quality dataset with about 43 QA pairs per video.

\begin{figure*}[h]
    \centering
    \includegraphics[width=0.87\textwidth]{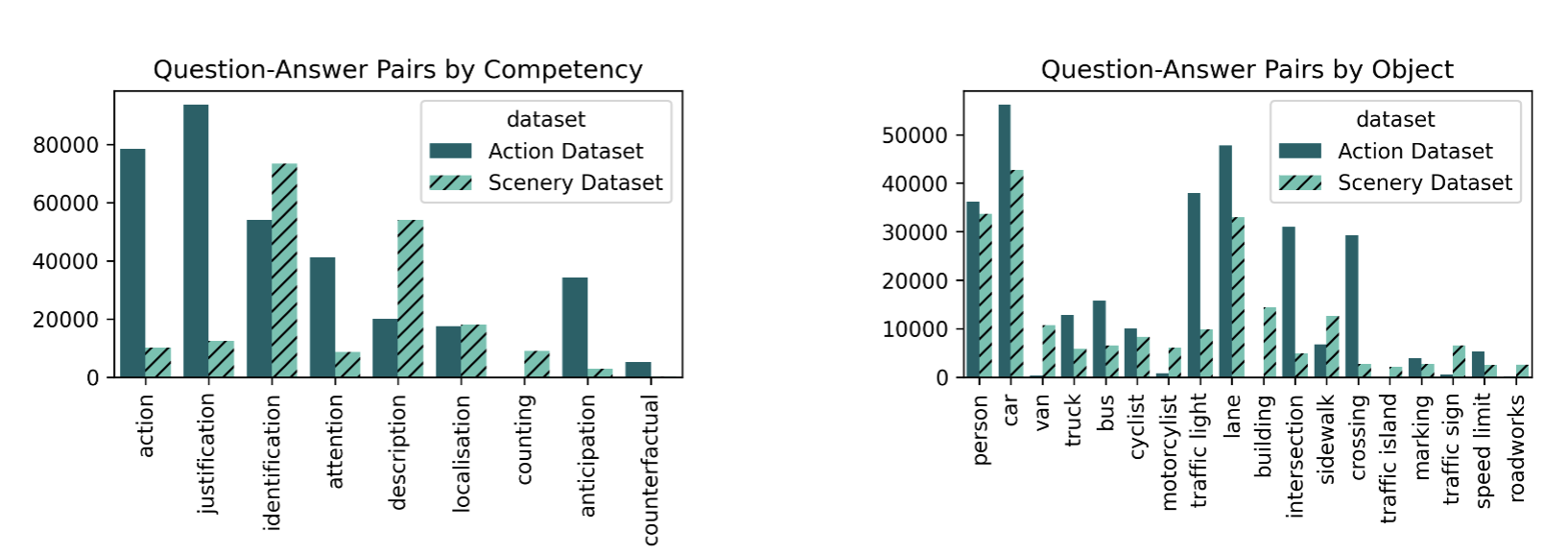}
    \caption{\textbf{Fine-tuning dataset statistics.} The \textit{Action} and \textit{Scenery} datasets have complementary strengths, one focused on action-justification competencies and one on description and localisation. The number of question-answer (QA) pairs for the competencies covered and for the objects referred are shown. One QA pair might cover more than one competency or object, hence the total is higher than dataset size.}
    \label{fig:dataset_stats}
\end{figure*}

\paragraph{Dataset statistics.} Our training dataset covers 9 different competencies: action (what the vehicle is doing), justification (why the action is taken), attention (what should be paid attention to in the current situation), identification (identifying an object given its description), localisation, description, counting, anticipation and reasoning given counterfactuals. The questions also cover a diverse set of objects, such as pedestrians, vehicles, cyclists, buildings, road infrastructure, signs, markings.
In Figure \ref{fig:dataset_stats}, we present the number of question and answer pairs for each of the 9 competencies above, as well as for the referred objects, for our two datasets, namely Action and Scenery. The complementary strengths of the datasets are apparent, with one focused on driving behaviours and one on perception tasks.

\begin{table}[tb]
  \centering
  \caption{\textbf{Dataset split.} The \textit{Action} dataset focuses on questions related to driving behaviours, the \textit{Scenery} dataset focuses on perception capabilities, while the evaluation dataset is designed to probe a range of competencies.} 
  \begin{tabular}{c}
    \begin{subtable}[b]{0.57\textwidth}
        \resizebox{\linewidth}{!}{
        \begin{tabular}{|l|c|c|c|}
            \hline & \textbf{Scenarios} & \textbf{QA pairs} & \textbf{QA per scenario} \\
            \hline
            Action & 24.5k & 267.8k & $\approx 10.9$ \\
            Scenery & 3.5k & 152.5k & $\approx 43.6$ \\
            Eval. Dataset & 100 & 1000 & $10$ \\
        \hline
        \end{tabular}}
    \end{subtable}
  \end{tabular}
  \label{tab:action_scenery}
\end{table}

\paragraph{Evaluation dataset.} \label{subsec:evaluation_dataset}
For evaluation we collected a small, low-density dataset from in-house human labelers, creating both the questions and the answers associated with the short videos. We labeled a small portion of held-out data on 500 human-generated questions using 20+ different evaluators to obtain our test set. Since answers are subjective and noisy, we labeled them twice, making sure the same evaluator does not receive the same question twice. After that, we manually reviewed the answers for semantic disagreements and mistakes. We relabeled such samples two more times and fixed the disagreements, preferring the semantics of the majority of responders but preserving maximal variety in the responses. Finally, we condensed this into 1k high-quality answers to 500 questions, with two correct but diverse answers per question, as shown in Table \ref{tab:action_scenery}. The dataset evaluates a range of competencies, including action and justification, attention, description, localisation, identification, counting and anticipation, as shown in Appendix \ref{appendix:dataset_examples}.

\section{Model Methodology}\label{sec:model}

We propose LingoQA Baseline, a vision language model for autonomous driving based on Vicuna v1.5 \cite{vicuna2023} with 7B parameters that can answer reasoning questions grounded in video outputs. We train a model that consumes a short video segment and produces answers to autonomous driving-related questions.

\begin{figure}[t]
\centering
\includegraphics[width=0.55\textwidth]{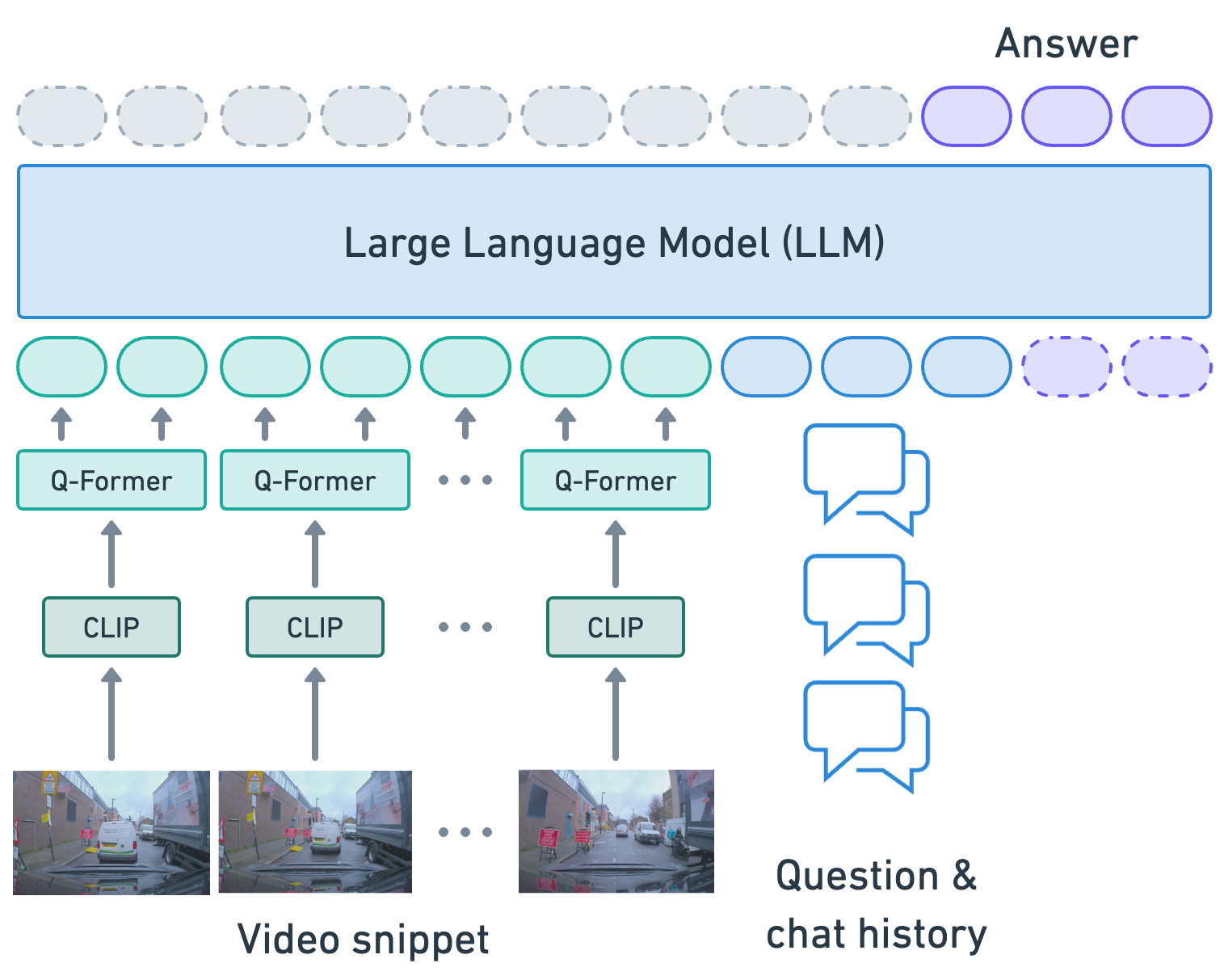}
\caption{\textbf{LingoQA Baseline model architecture.} We first encode individual frames using CLIP and Q-Former. The Q-Former outputs tokens and we feed the tokens from all frames along with chat history and questions into the LLM, which then predicts an answer.}
\label{fig:architecture}
\end{figure} 

\subsection{Architecture}

The LingoQA Baseline model architecture is based on recent VLMs \cite{blip2,liu2023visual,driess2023palme} but enhances them by incorporating a video encoding strategy to process multiple frames from a video snippet, as shown in Figure \ref{fig:architecture}. 

\paragraph{Vision encoder.} 
We use CLIP \cite{radford2021learning}, a Vision Transformer (ViT) pre-trained contrastively on image-language pairs, to encode images into features. The inputs to the vision encoder are RGB images from the front camera.  We squash the input images to a size of $224\times224$ as opposed to cropping them in order to keep the full image context. Subsequently, we pass the features through a transformer network, the Querying Transformer (Q-Former), that akin to BLIP-2 \cite{blip2} acts as a bridge between the vision and language feature spaces. The embeddings are then projected into the large language model (LLM) space using a linear projection layer. We repeat this process for $T = 5$ frames of the input video and concatenate the tokens from each image.

\paragraph{Large language model.} 
We leverage pretrained LLMs to give LingoQA Baseline the ability to answer general questions related to both driving scenes, as well as general knowledge. We use Vicuna v1.5 \cite{vicuna2023} with 7B parameters built on top of Llama-2 \cite{touvron2023llama}. The language model is autoregressive and hence can be conditioned on textual inputs, as well as image tokens. The training objective is to predict the next language token in a sequence. We mask all tokens from the training loss that belong to the text prompt, including question and chat history. 

\subsection{Training Recipe} 
Our training uses a two-step approach to better utilise video features and improve learning when answering questions based on video data. Through this two-step training, we aim to better understand and use video data. In the first stage, we train the self-attention layers for the LLM and the vision encoder (QKV), together with all the Q-Former parameters and the linear language projection layer. In the second stage, we fine-tune the same parameters as in the previous stage, keeping the vision encoder frozen. Further details regarding training parameters are presented in Appendix \ref{training:parameters}.

\label{subsec:training_recipe}
\paragraph{Stage 1: Pre-training for feature alignment.}
In the first stage, we pre-train the self-attention layers of the LLM and the vision encoder (QKV) on the GQA and SVIT datasets to align image features with the embedding space of the pretrained LLM. The GQA dataset \cite{hudson2019gqa} contains more than 22M questions over 113k images. The recently introduced SVIT dataset \cite{zhao2023svit} contains 4.2M question-answer pairs over 108.1k images. We leverage initial weights from different models to accelerate the training process. We initialise the vision encoder using publicly available weights of OpenCLIP \cite{ilharco_gabriel_2021_5143773}, the Q-Former from BLIP2 weights \cite{li2023blip2}, and language model from Vicuna v1.5 \cite{vicuna2023}.

\paragraph{Stage 2: Fine-tuning for video QA.}
In the second stage, we fine-tune the model on our video question-answering Action and Scenery datasets described in Section \ref{subsec:dataset}. During the fine-tuning phase, each sample is composed of 5 frames taken from a 4-second span of video, accompanied by a QA-pair. The dataset used to fine-tune LingoQA Baseline is open-sourced.

\section{Empirical Evaluation on LingoQA} \label{sec:evaluation_on_lingoqa} 
\label{sec:empirical_evaluation}

\begin{table*}[t]
\centering
\caption{\textbf{Empirical evaluation on LingoQA.} Ablation study highlighting the impact of various modifications in training recipes, dataset composition, frame count, video processing techniques and language model.}
 \resizebox{\linewidth}{!}{%
\begin{tabular}{llcccccc}
\toprule
& Ablation & Lingo-Judge [\%] $\uparrow$ & BLEU $\uparrow$ & METEOR $\uparrow$  & CIDEr $\uparrow$ \\
\midrule
& \textbf{LingoQA Baseline} & $\textbf{60.80}$ & $\textbf{15.00}$ & $18.56$ & $\textbf{65.61}$ \\
\midrule
\multirow{2}{5cm}{\textbf{\textit{Training recipe}}\\Instead of pre-train and fine-tune} 
& No fine-tuning & $33.60$ & $8.33$ & $14.33$ & $39.16$ \\ 
& No pre-training & $56.60$ & $13.53$ & $17.91$ & $57.98$  \\
\midrule
\multirow{2}{5cm}{\textbf{\textit{Fine-tuning dataset}}\\Instead of action and scenery} 
& Action only & $53.80$ & $11.65$ & $17.68$ & $46.50$ \\
& Scenery only & $55.40$ & $13.00$ & $18.38$ & $55.88$ \\
\midrule 
\multirow{3}{5cm}{\textbf{\textit{Frame count}}\\Instead of 5 frames}
& Single frame & $57.00$ & $14.21$ & $18.40$ & $59.46$ \\
& 3 frames & $59.80$ & $14.61$ & $18.44$ & $62.61$ \\
& 7 frames & $60.60$ & $14.46$ & $\textbf{18.61}$ & $61.82$ \\ 
\midrule 
\multirow{2}{5cm}{\textbf{\textit{Video fusion}}\\Instead of late-fusion}
& Early-fusion & $48.40$ & $13.98$ & $17.61$ & $61.42$ \\
& Mid-fusion & $59.20$ & $14.44$ & $18.47$ & $63.05$\\
\midrule
\multirow{3}{5cm}{\textbf{\textit{Language model}}\\Instead of Vicuna-1.5-7B\cite{vicuna2023}}
& OPT-7B \cite{zhang2022opt} & $50.00$ & $14.98$ & $15.99$ & $60.08$ \\
& Llama-2-7B-Chat \cite{touvron2023llama2} & $59.20$ & $13.52$ & $18.43$ & $59.87$ \\
& Mistral-7B-Instruct \cite{jiang2023mistral} & $58.00$ & $13.80$ & $18.33$ & $64.21$ \\
\bottomrule
\end{tabular}}
\label{ablation_table_models}
\end{table*}

\subsection{Ablation Studies on LingoQA}
With the highly modular architecture of VLMs, the question remains what architectural components of the LingoQA Baseline model and dataset composition contribute the most to its performance? We conduct several ablation studies around the architecture and training paradigm described in Section \ref{sec:model}. We investigate variations to the \textit{training strategy}, \textit{training data composition}, \textit{frame count}, \textit{video fusion methods}, and the use of different \textit{large language models}, as shown in Table \ref{ablation_table_models}. The results are obtained by having each model generate one answer per question and then compare the predicted answer to the two ground truth answers. Examples of comparisons between our baseline model's answers and answers from other models from the ablations are presented in Appendix \ref{model:examples}.

\paragraph{Training Recipe and Dataset Mixture.} 
The aim of the training strategy experiments is to understand how much the pre-training and the fine-tuning steps contribute to performance. Fine-tuning on the LingoQA dataset doubles the performance over generic VQA pre-training. The \textit{Action Dataset} and the \textit{Scenery Dataset} both prove influential in improving model performance.

\paragraph{Impact of Frame Count.} We investigated the variation in VQA performance with decreasing and increasing the number of video frames fed into the model. The base model contains 5 frames over a 4-second context. The performance declined when shifting from multi-frame video to a single image representation, while remaining close to the multi-frame baseline, showing that a certain proportion of autonomous driving scenarios can be solved from a single frame, as further discussed in Section \ref{sec:sota_eval}. Nonetheless, to fully reach human-level multi-frame performance, video fusion is needed, as shown in Table \ref{tab:sota_evaluation}. We conclude that both improved single-frame reasoning and video fusion are required.

\paragraph{Impact of Video Fusion Strategy} This study explores three methods for integrating video frames into the LLM: \textit{early-fusion}, \textit{mid-fusion}, and \textit{late-fusion}. The \textit{early-fusion} method employs average pooling to condense features from the vision encoder prior to their incorporation into the Q-Former, producing a unified visual feature vector for language space projection. The \textit{mid-fusion} approach, merges video features into fixed-size tokens within the Q-Former with the cross-attention mechanism. The \textit{late-fusion} method feeds individual frame embeddings from the Q-Former output into the LLM. Our findings demonstrate that both \textit{mid-fusion} and \textit{late-fusion} are effective methods for incorporating video content into the model.

\paragraph{Impact of Large Language Model}
We investigate the impact that different Large Language Models have on the overall performance of our vision-language model. The best score is achieved by Vicuna-1.5-7B \cite{vicuna2023}. In the same family of models, Llama-2-7B \cite{touvron2023llama2} achieves comparable, but slightly lower performance. Despite the promise of improved performance, Mistral-7B \cite{jiang2023mistral} is less effective in our fine-tuning task. OPT-7B \cite{zhang2022opt} substantially underperforms the others, potentially due to the lower embedding size - 1048 compared to 4096 for all other models.

\subsection{Evaluation of SOTA Vision-Language Models}
\label{sec:sota_eval}
To demonstrate the relevance of the newly proposed benchmark, we evaluate a series of SOTA vision-language models and compare them to human performance, as shown in Table \ref{tab:sota_evaluation}. 

\paragraph{Human study.} Human performance is evaluated on both video inputs and a single frame. We find a performance degradation from 96.6\% to 81.8\% without temporal context. The main failure modes include misclassifying parked cars as engaged in traffic,  miscounting developing hazards such as motorcyclists and pedestrians over multiple frames, missing state transitions of traffic lights, and
failing to predict the correct speed when behind a vehicle. These results indicate that while video understanding is required for reaching human multi-frame performance, improvements in single-frame reasoning are also crucial.

\paragraph{Fine-tuned models.} Our work identifies a notable 23\% performance gap between single-frame LLaVA and single-frame human capability, underlining
the benchmark’s significance for the autonomous driving and vision-language community. Our evaluation includes fine-tuning a single-frame LLaVA, a single-frame BLIP-2, and a text-only Vicuna-7B model on LingoQA, as shown in Table \ref{tab:sota_evaluation}. Results reveal a performance lower bound of 38\% for models lacking visual inputs. Notably, vision-language models, such as BLIP-2 and LLaVA, surpass this text-only baseline by 13\% and 20.2\%, respectively, with performance enhancements attributed to enhanced perceptual capabilities. Furthermore, LLaVA’s use of a larger CLIP crop size (336) compared to 224 improves performance.

\paragraph{Zero-shot models.} The best zero-shot model GPT-4V performance is still 37\% below the multi-frame human performance. This highlights that further advancements are required for frontier models to fully solve the benchmark. The Lingo-Judge accuracy is impacted by the model response style, where long-form incorrect answers may receive high ratings, as it happens with FUYU. Further details are presented in Appendix \ref{appendix:generalisation}.

\begin{table}
\centering
\caption{\textbf{Evaluating vision-language models on LingoQA}. The performance of existing vision-language models is far from human capability.}
\resizebox{0.65\linewidth}{!}{
\begin{tabular}{@{}l|c|c|ccccc@{}}
\toprule
 & Category & No. Frames & Human & Lingo-J & BLEU & METEOR & CIDEr  \\
\midrule
 \textbf{Human} & \multirow{2}{3cm}{\textit{  human study}} & 5 & \textbf{93.3} & \textbf{96.6} & \textbf{81.04} & \textbf{52.92} & \textbf{361.77} \\
 Human & & 1 & - & 81.8 & 10.64 & 15.01 & 64.45 \\
\midrule
 \textbf{LingoQA} & \multirow{5}{3cm}{\textit{  fine-tuned models}} & 5 & \textbf{57.1}  & \textbf{60.8} & \textbf{15.00} & \textbf{18.56} & \textbf{65.62} \\
LingoQA & & 1 & - & 57.0 & 14.21 & 18.40 & 59.46 \\
LLaVA & & 1 & - & 59.0 & 12.5 & 18.5 & 57.0 \\
BLIP-2 & & 1 & - & 52.2 & 13.0 & 17.4 & 60.1 \\
Vicuna-7B & & 0 & - & 38.8 & 10.1 & 15.2 & 51.0 \\
\midrule
\textbf{GPT-4V} & \multirow{4}{3cm}{\textit{  zero-shot models}} & 5 & \textbf{56.61} & \textbf{59.6} & 6.30 & 12.35 & \textbf{42.82} \\
LingoQA & & 5 & - & 33.6 & \textbf{8.33} & \textbf{14.33} & 39.16 \\
LLaVA & & 1 & 38.97 & 49.4 & 4.23 &  8.38 &  38.39 \\
FUYU & & 1 & 17.69 & 45.4 & 1.90 & 13.00 & 12.04 \\
\bottomrule
\end{tabular}}
\label{tab:sota_evaluation}
\end{table}

\section{Discussion and Limitations}

\paragraph{Strengths of Lingo-Judge.} 
The strength of our contribution comprises proposing a classifier that is \textit{highly correlated} with human inputs and \textit{efficient} to run. In conjunction with the evaluation dataset that we propose, it becomes a useful tool for benchmarking vision-language models for autonomous driving for visual question answering task, which has been historically challenging to evaluate in a consistent fashion. With this contribution, autonomous driving research can be accelerated by providing a reliable, efficient, and easy-to-interpret benchmark.

\paragraph{Limitations of Lingo-Judge.} The Lingo-Judge is specifically tailored
for open-vocabulary evaluation on the LingoQA benchmark. While it demonstrates high speed and accuracy compared to larger models like GPT-4, it is designed akin to the model described in the TruthfulQA \cite{lin2022truthfulqa} paper, which suggests that such specialized models are not expected to generalize well to new questions. Second, we optimized the classifier to evaluate responses in the style provided by human annotators in the evaluation dataset. The same response style is adopted in the LingoQA training sets and the models. Further details regarding generalisation to response styles is studied in Appendix \ref{appendix:generalisation}. Third, as the classifier is only trained to predict factual correctness, it cannot discern which answer of two equally correct answers humans prefer. 

\paragraph{Dataset and model limitations.} 
One of the primary constraints is that our model operates on relatively short video segments and few frames, limiting the contextual understanding of scenarios. We also do not test for driving decisions and attention mechanisms, focusing on question-answering abilities only. We did not test the scaling in our models and focused on the most practical 7B parameter LLMs only. Our dataset and baseline are limited to information from a single front-facing car camera, excluding additional sensory inputs like LiDAR that could enrich the model's understanding of the driving environment. Expanding the model to address the short video context, as well as adding action prediction and evaluation to the dataset and the benchmark would result in a more versatile system for autonomous driving.

\section{Conclusion}
In this paper, we introduced a novel benchmark for Visual Question Answering for autonomous driving. The benchmark consists of an evaluation dataset, learned classifier-based metric Lingo-Judge that is highly correlated with human evaluation, a comprehensive high-quality training dataset for autonomous driving. The fast feedback from employing Lingo-Judge facilitates effective exploration in the visual QA field. Additionally, the comprehensive experiments on different model combinations presented in this paper can become a foundation for further improvement of end-to-end autonomous driving systems. The LingoQA benchmark is openly released to spur further community research, providing a reliable and highly correlated evaluation method to human ratings. 

\section*{Acknowledgements}
This work was possible through the help of many colleagues across Wayve. In particular we would like to acknowledge the support from: Anthony Hu, Miguel Sanchez Lohff, Lorenzo
Bertoni, Charlie Lyons-Rothbart, Emma Wang, Harriett-Rose Follas, Kyle Esecson, Ben Foxall, Naama Zahavi,
Ruben Diaz, Rudi Rankin, Tilly Pielichaty, Sarah Belghiti,
Giulio D’Ippolito, Dan Reisman, Alex Persin, Fergal Cotter, Przemyslaw Mazur, Thomas Sajot, Giacomo Gallino,
Alex Garcia Mayans, Tim Geypens, Robin Tweedie, Rebecca Hills.

%
%
\bibliographystyle{splncs04}
\bibliography{main}

\appendix

\clearpage

\section{LingoQA Dataset Examples}\label{appendix:dataset_examples}
\begin{figure*}[h!]
\centering
\includegraphics[width=\textwidth]{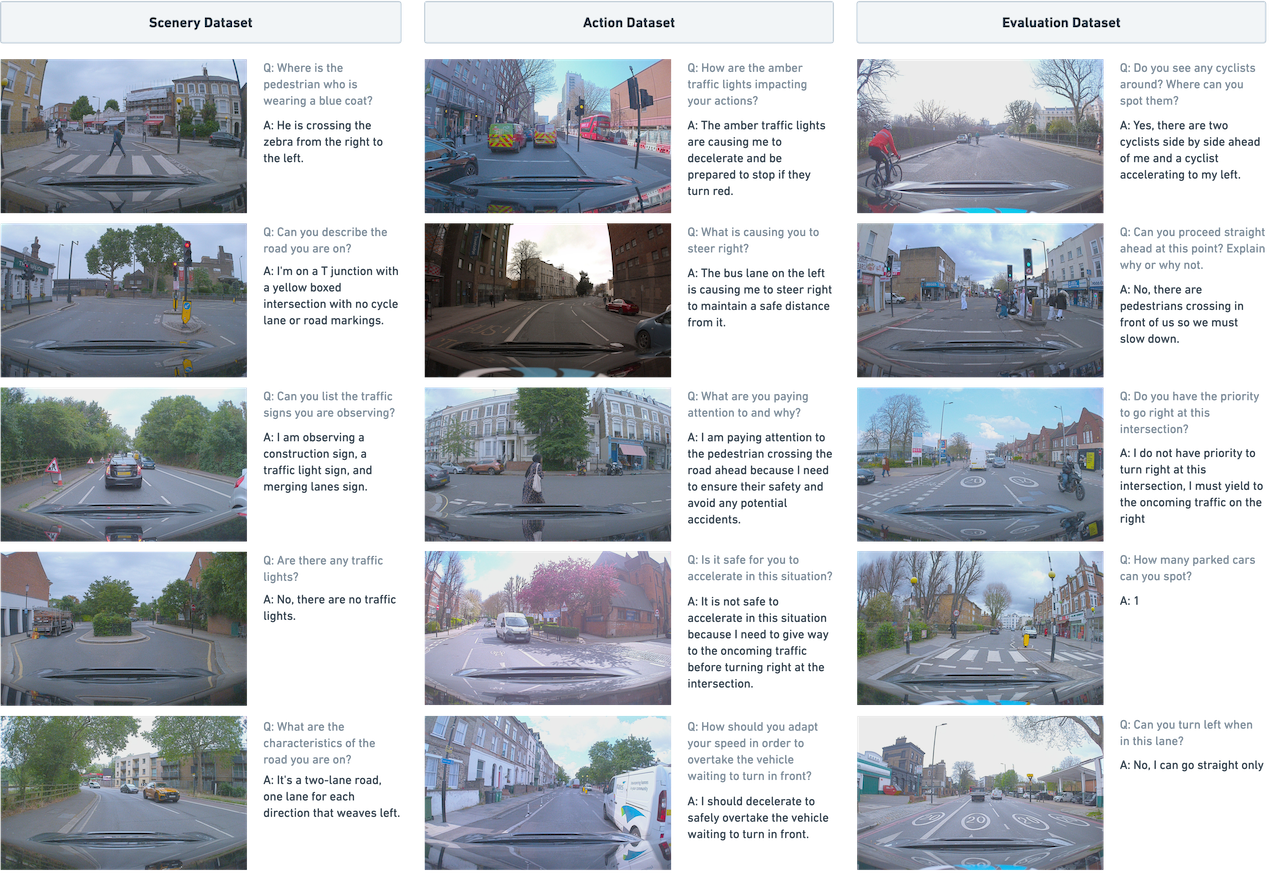}
\caption{\textbf{LingoQA dataset examples.} From left to right: \textit{scenery} dataset, \textit{action} dataset, and \textit{evaluation} dataset. Further video examples are provided in the supplementary material accompanying the submission.}
\label{fig:more_dataset_samples}
\end{figure*}

\begin{figure*}[h!]
\centering
\includegraphics[width=\textwidth]{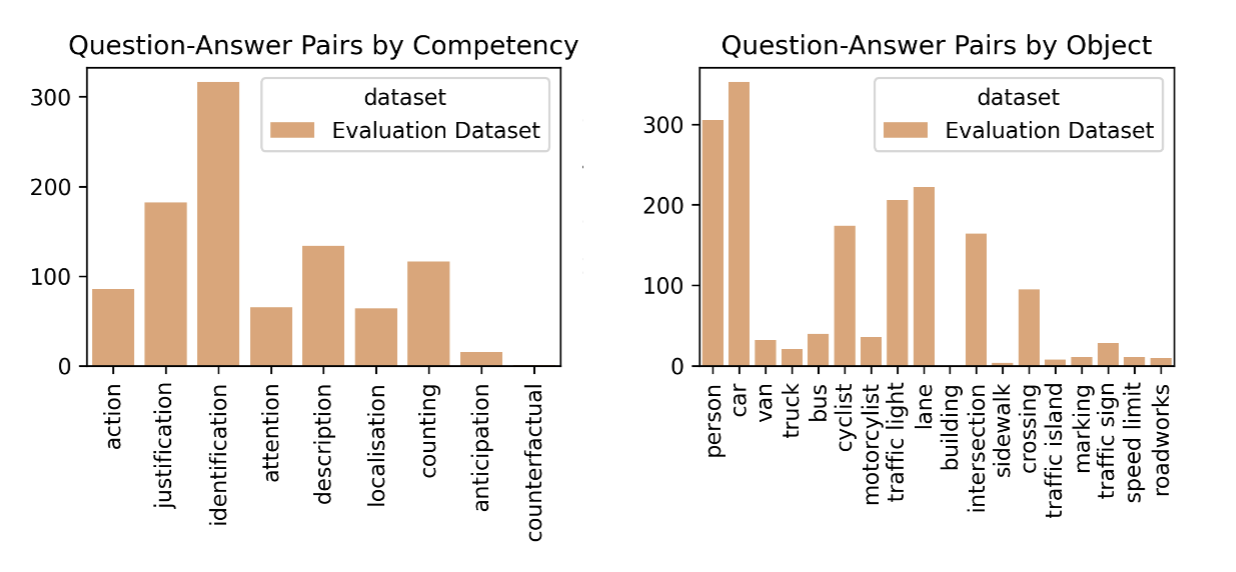}
\caption{\textbf{Evaluation Dataset Statistics.} The evaluation dataset assesses a range of competencies and includes a wide range of objects relevant for autonomous driving.}
\label{fig:eval_dataset_statistics}
\end{figure*}

Further examples on the capabilties existent in the training and the evaluation datasets are shown in Figure \ref{fig:more_dataset_samples}. The \textit{scenery} dataset contains highly descriptive elements, such as object colours, junction type, construction zones, traffic lights, and the road layout. The \textit{action} dataset is complementary and focused on driving competencies, such as the impact of traffic lights on driving and interactions with other road agents. The \textit{evaluation} dataset contains a broad range of questions aimed to test competencies relevant for autonomous driving. Further examples from the evaluation benchmark are also included in the overview video accompanying the submission. The dataset statistics for the evaluation dataset are shown in Figure \ref{fig:eval_dataset_statistics}. Notably, any personal identifiable information, such as faces and plate ID's, has been anonnymised in the dataset.

\section{Lingo-Judge Examples}\label{classifier:examples}
We present additional qualitative examples from our evaluation dataset in Table \ref{tab:metrics_comparison}, alongside predictions from our base model and corresponding metrics for each individual sample. Metrics based on \textit{n-gram} matching such as CIDEr tend to be error-prone. For example, expressions that have the same meaning, but entirely different words, are marked as not similar at all, such as \textit{``None''} and \textit{``There are no cars.''}. Sentences with minor but significant differences are graded as highly similar, despite having opposite meanings, such as \textit{``The traffic lights are showing green''} and \textit{``The traffic lights are showing red''}. Lingo-Judge demonstrates robustness against these varied expressions and subtle changes. Lingo-Judge also has limitations, primarily seen when establishing the correctness of the answer would require extra context from the videos. These examples can be seen in Table \ref{tab:failure_cases}.

\begin{figure*}
\centering
\includegraphics[width=0.85\textwidth]{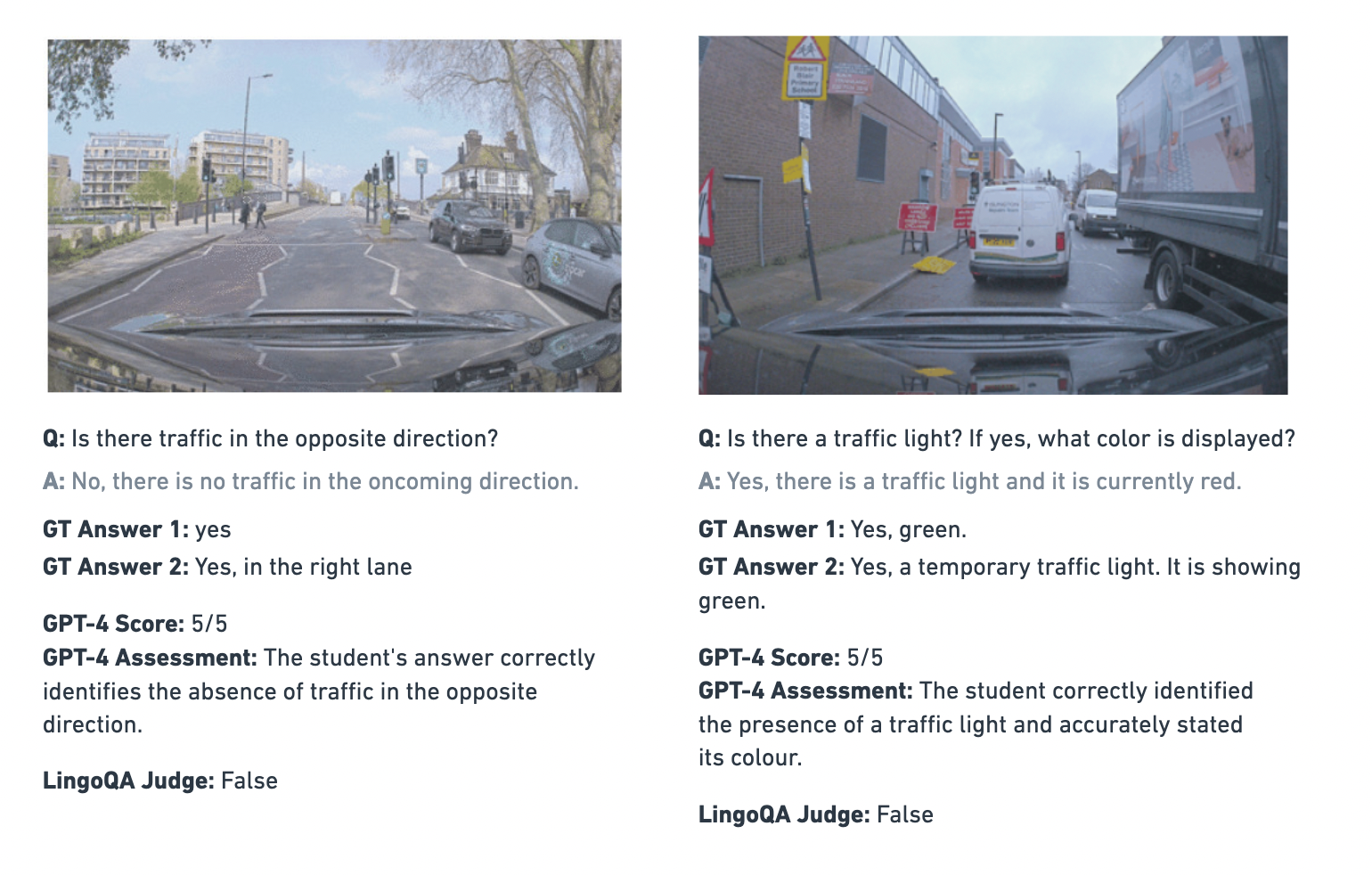}
\caption{\textbf{Classifier examples.} Examples of Lingo-Judge outputs compared to GPT-4.}
\label{fig:classifier_examples}
\end{figure*}

We qualitatively compare our classifier to GPT-4 ratings. These examples are shown in Figure \ref{fig:classifier_examples}. In this situation, GPT-4 is misled by the fact that the model answer contains partially correct information. The GPT-4 assessment states that \textit{``The student correctly identified the presence of a traffic light''} and, despite the colours not being correct, further explains that \textit{``and accurately stated its colour''}. This highlights some challenges faced by GPT-4 when trying to rate the truthfulness of an answer. Lingo-Judge correctly identifies that the statements described by the model are false.

\begin{table*}[hbt]
  \small 
  \setlength{\tabcolsep}{3pt} 
  \centering
    \caption{\textbf{Qualitative comparison of metrics.} Questions and labels from our evaluation dataset along with exemplary predictions and their corresponding sample-level metrics. Notable is that the classifier (here ``L-J'') can pick up slight but crucial differences between ground truth and prediction to correctly identify a wrong prediction, such as ``The traffic lights are green'' and ``The traffic lights are red'', in which case CIDEr still shows high agreement. \textcolor{green}{Green} color indicates agreement with human judgement, while \textcolor{orange}{orange} and \textcolor{red}{red} show disagreement.}
  \resizebox{\linewidth}{!}{
  \begin{tabular}{
    >{\raggedright\arraybackslash}p{3cm} 
    >{\raggedright\arraybackslash}p{2.5cm} 
    >{\raggedright\arraybackslash}p{2.5cm} 
    ccccccccc}
  \toprule
  Question & Label & Prediction  & CIDEr & GPT4 & GPT4-CoT & L-J Prob. & L-J Class. \\ 
  \thinmidrule
  How many pedestrians are crossing the road? & Zero pedestrians & There are no pedestrians crossing the road.  & \textcolor{red}{23.18}  & \textcolor{green}{5} & \textcolor{green}{5} & 0.96 & \textcolor{green}{True} \\
  \thinmidrule
  What is the road speed limit? & 20 mph - it is written on the road & The road speed limit is 20 mph.  & \textcolor{orange}{97.91}  & \textcolor{green}{4} & \textcolor{orange}{3}  & 0.95 & \textcolor{green}{True} \\
  \thinmidrule
  How many cars are driving in your direction? & None & There are no cars driving in my direction.  & \textcolor{red}{0.11} & \textcolor{green}{4} & \textcolor{orange}{3} & 0.96 & \textcolor{green}{True} \\
  \thinmidrule
  Which vehicle should you follow if any? & The motorcyclist. & If any, I should follow the motorcycle ahead. & \textcolor{red}{0.42} & \textcolor{green}{4} & \textcolor{green}{5} & 0.95 & \textcolor{green}{True} \\
  \thinmidrule
  What is the current action and its justification? Answer in the form ``action, justification'' & Slow down, there is a stationary van infront of us & I am decelerating because of the stationary truck ahead. & \textcolor{red}{34.79} & \textcolor{green}{5} & \textcolor{green}{5} & 0.96 & \textcolor{green}{True} \\
  \thinmidrule
  What is the current action and its justification? Answer in the form ``action, justification'' & Stop, Red light & I am stopping because the traffic lights to go straight are red. & \textcolor{red}{18.62}  & \textcolor{green}{5} & \textcolor{green}{5} & 0.95 & \textcolor{green}{True} \\
  \thinmidrule
  How many cyclists can you see? & I can see 3 cyclists & I can see two cyclists. & \textcolor{orange}{150.29} & \textcolor{green}{1} & \textcolor{orange}{2} & 0.05 & \textcolor{green}{False} \\
  \thinmidrule
  What color are the traffic lights showing? & The traffic lights are showing green & The traffic lights are showing red. & \textcolor{red}{329.36} & \textcolor{green}{0} & \textcolor{green}{1} & 0.05 & \textcolor{green}{False} \\
  \thinmidrule
  What action are you taking with respect to the cyclist? & Overtaking them on the right and keeping the speed & I am overtaking the cyclist on the left.  & \textcolor{red}{349.52} & \textcolor{orange}{2} & \textcolor{orange}{2} & 0.10 & \textcolor{green}{False} \\
  \thinmidrule
  In which direction is the bus driving? & The bus is driving in the opposite direction & The bus is driving in the oncoming direction.  & \textcolor{red}{404.65} & \textcolor{red}{4} & \textcolor{red}{5} & 0.31 & \textcolor{green}{False} \\
  \thinmidrule
  Are there any parked car on the side of the road? & Yes, there are two cars parked on the right of the road	& No, there are no parked cars on either side of the road. & \textcolor{orange}{142.40} & \textcolor{green}{0} & \textcolor{green}{0} & 0.05 & \textcolor{green}{False}\\
  \thinmidrule
  Is acceleration necessary in this situation? If so, provide the reason. & No. We should decelerate in this situation because there is a vehicle stopping ahead of us. & No, acceleration is not necessary in this situation as I am already driving at the speed limit.  & \textcolor{red}{177.96} & \textcolor{orange}{3} & \textcolor{orange}{3} & 0.31 & \textcolor{green}{False} \\
  \bottomrule
  \end{tabular}}
  \label{tab:metrics_comparison}
  \end{table*}

  \begin{table*}[hbt]
  \small 
  \setlength{\tabcolsep}{3pt} 
  \centering
   \caption{\textbf{Failure Cases of Lingo-Judge.} Examples where Lingo-Judge makes a wrong judgement about the correctness of the model prediction. \textcolor{green}{Green} color indicates agreement with human judgement, while \textcolor{orange}{orange} and \textcolor{red}{red} show disagreement.} 
  \resizebox{\linewidth}{!}{
  \begin{tabular}{
    >{\raggedright\arraybackslash}p{3cm} 
    >{\raggedright\arraybackslash}p{2.5cm} 
    >{\raggedright\arraybackslash}p{2.5cm} 
    cccccccc}
  \toprule
  Question & Label & Prediction & CIDEr & GPT4 & GPT4-CoT & L-J Prob. & L-J Class. \\ 
  \thinmidrule
  Is any vehicle stopped at the zebra crossing? & There is no zebra crossing. & No, there are no vehicles stopped at the zebra crossing. & \textcolor{red}{106.19} & \textcolor{red}{4}    & \textcolor{green}{2} & 0.93 & \textcolor{red}{True} \\
  \thinmidrule
  What elements are capturing your attention and what is the reason behind it? & I am making sure there is no vehicle on the left lane I am making my lane change to. I am also looking at the traffic light to make sure it stays green while I am crossing the intersection. & I am paying attention to the bus lane on the left because I need to maintain a safe distance from it. & \textcolor{green}{0.91} & \textcolor{orange}{3} & \textcolor{green}{1} & 0.57 & \textcolor{red}{True} \\
  \thinmidrule
  What is the current action and its justification? Answer in the form "action, justification" & We are taking a left turn, because we are at a t-junction. Then we keep a safe distance from the cyclists in front. & I am decelerating to keep a safe distance from the cyclist ahead of me. & \textcolor{orange}{46.45} & \textcolor{orange}{2} & \textcolor{orange}{2} & 0.32 & \textcolor{red}{False} \\
  \bottomrule
  \end{tabular}}
  \label{tab:failure_cases}
  \end{table*}

\section{GPT-4 Grading}\label{appendix:chatgpt}
In this section we provide an overview of the implementation details for the evaluation method using GPT-4 with and without Chain-of-Thought (CoT) \cite{wei2023chainofthought} prompting.  

\begin{figure*}
\centering
\includegraphics[width=0.95\textwidth]{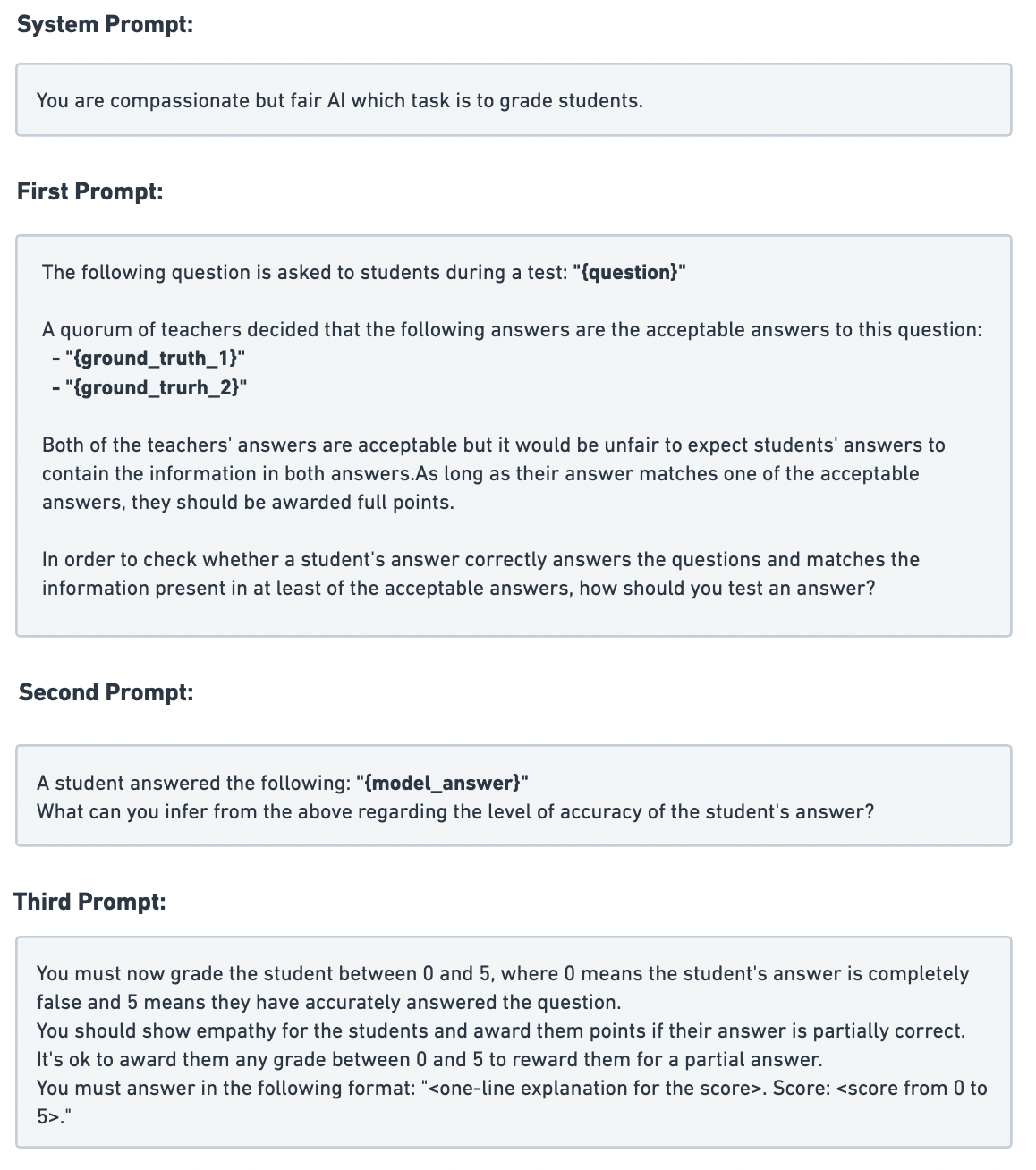}
\caption{\textbf{GPT-4 with Chain of Thought (CoT) prompting.} First, GPT-4 is provided with the question and ground truth answers, and asked to come up with a strategy for testing the answer. Second, GPT-4 is provided with the model answer and is prompted to evaluate the accuracy of the response based on the previously defined strategy. Finally, GPT-4 is asked to provide a grade for the student.}
\label{fig:gpt4_cot}
\end{figure*}

\textbf{GPT-4 with CoT.} In order to evaluate a model's answer with GPT-4 and CoT prompting, we first provide GPT-4 with the question and one or more valid answers for the questions, and ask it to come up with a strategy to evaluate new answers to this question. We then provide GPT-4 with the model's answer and ask it to evaluate the answer using the strategy it proposed in the previous step. Finally, we ask GPT-4 to give the model a grade between 0 and 5, where 5 means the answer is perfect. The prompt used is shown in Figure \ref{fig:gpt4_cot}. 

\begin{figure*}
\centering
\includegraphics[width=0.95\textwidth]{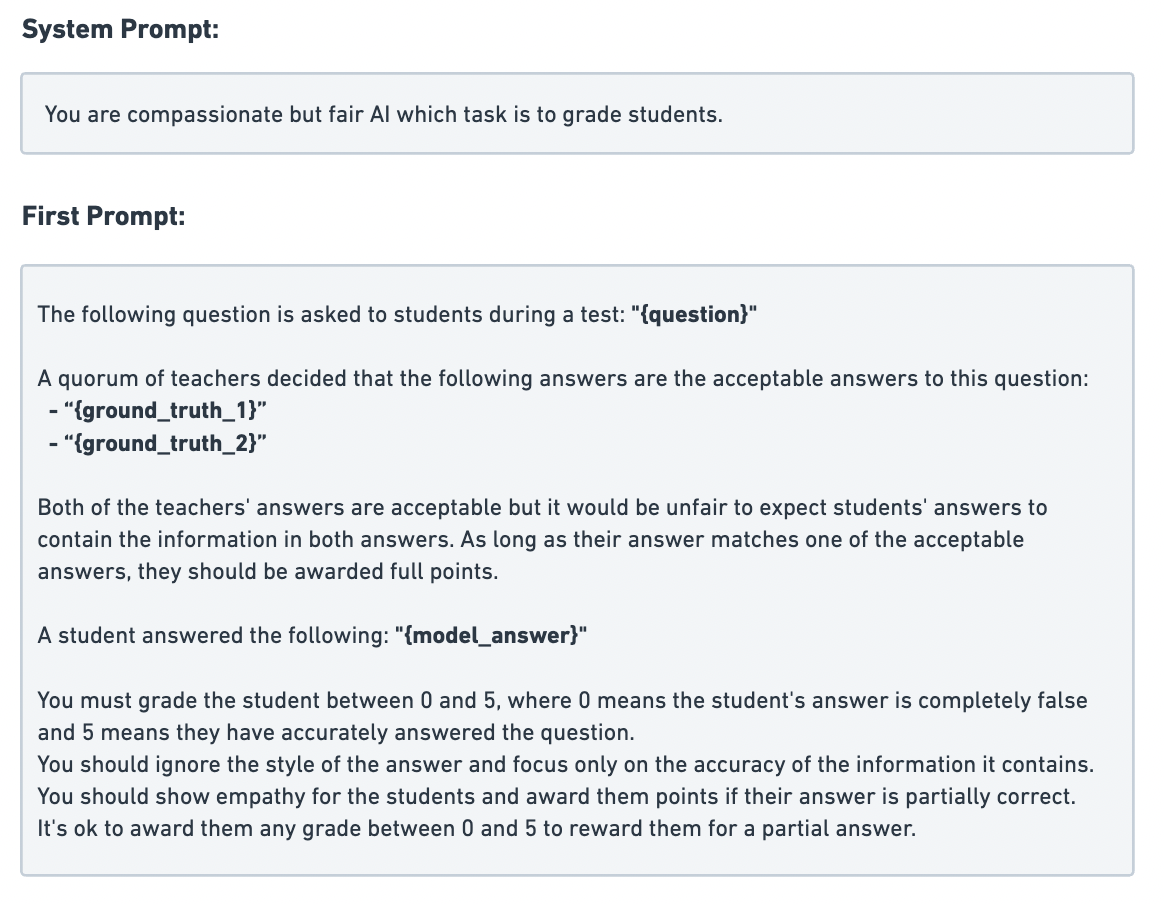}
\caption{\textbf{GPT-4 without Chain of Thought (CoT) prompting.} GPT-4 is provided with a prompt that contains the question, the ground truth answers, and the model response, and is requested to directly provide a grade for the student.}
\label{fig:gpt4_no_cot}
\end{figure*}

\textbf{GPT-4 without CoT.} When evaluating model outputs without CoT prompting, we provide GPT-4 with the question, one or more valid answers for the questions, and the model predictions and we directly ask GPT-4 to give the model a grade between 0 and 5, without the intermeidate reasoning steps. The prompt used is shown in Figure \ref{fig:gpt4_no_cot}.

We emit concurrent requests to our Azure's GPT-4 deployment in order to max-out the limit of 40k tokens per minute. GPT-4 without CoT prompting required more than 13 minutes to perform the evaluation, and GPT-4 with CoT prompting requires more than 50 minutes.

\section{Lingo-Judge Correlation Study}\label{correlation:study}

We show that Lingo-Judge exhibits a higher correlation to human judgment than commonly-used language-based metrics, and than GPT-4.
To do so, we computed the scores of 15 different models and 2 groups of human labellers on the questions in our evaluation dataset using Lingo-Judge, GPT-4, BLEU4, METEOR and CIDEr. These scores are reported in Table \ref{correlation_table_models}.

We then computed the Pearson and Spearman correlation coefficients between these metrics and the human evaluation.
The \textbf{Pearson correlation coefficient} measures the strength of the linear correlation between the human evaluation and a metric score, while the \textbf{Spearman rank correlation coefficient} measures the monotonic relationship between the human evaluation and the metric. The higher the Spearman coefficient, the better a metric is at ranking answers in the same order as our human evaluators.
To compute the confidence intervals, we use the Fisher transformation with a 95\% confidence level.

\begin{figure}
\centering
\includegraphics[width=0.95\textwidth]{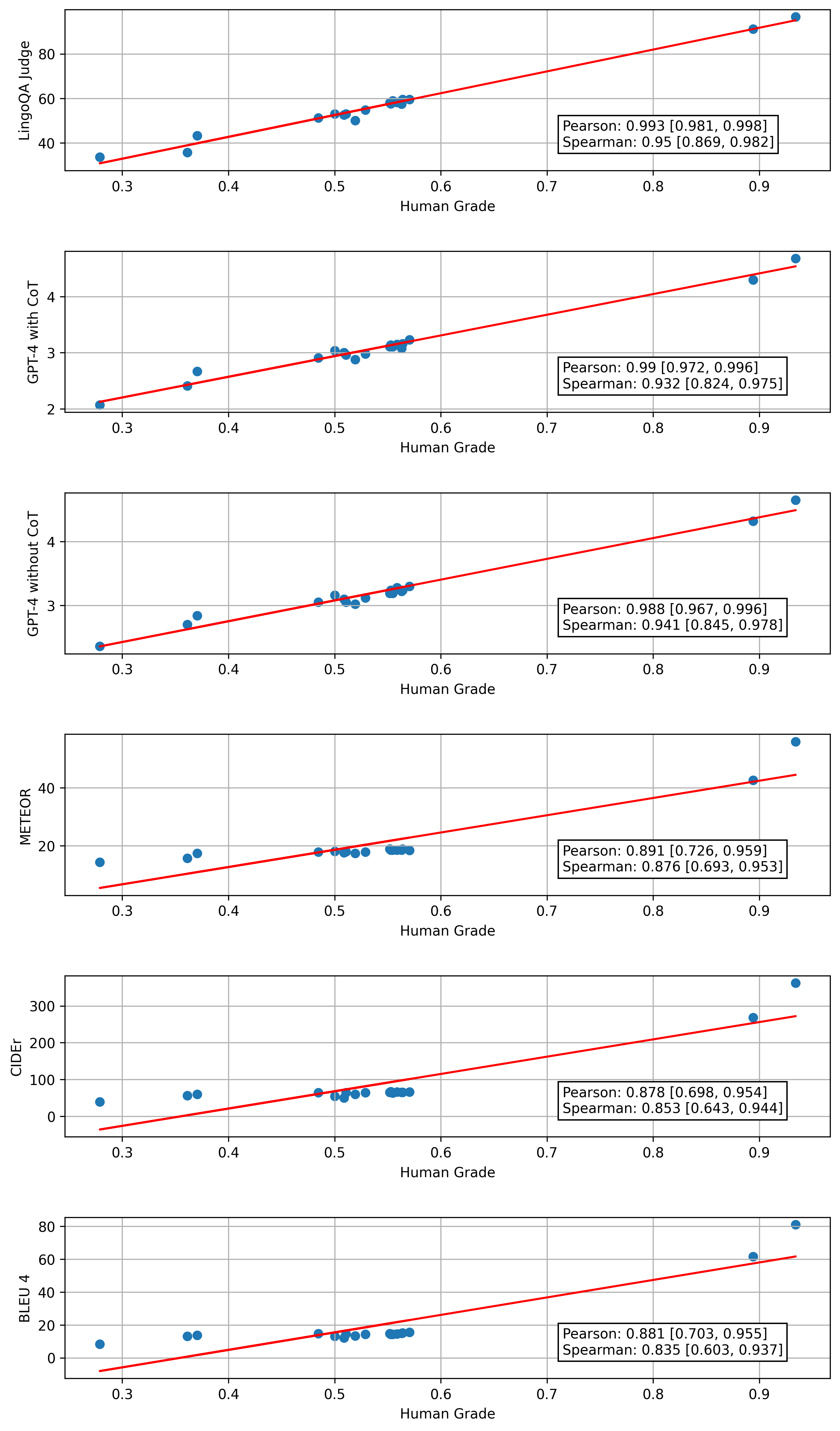}
\caption{\textbf{Correlation trends.} Correlation trends of the average grade of models compared to the average human-grades, for different metrics.}
\label{fig:correlation_trends}
\end{figure}

\begin{figure}
\centering
\includegraphics[width=0.95\textwidth]{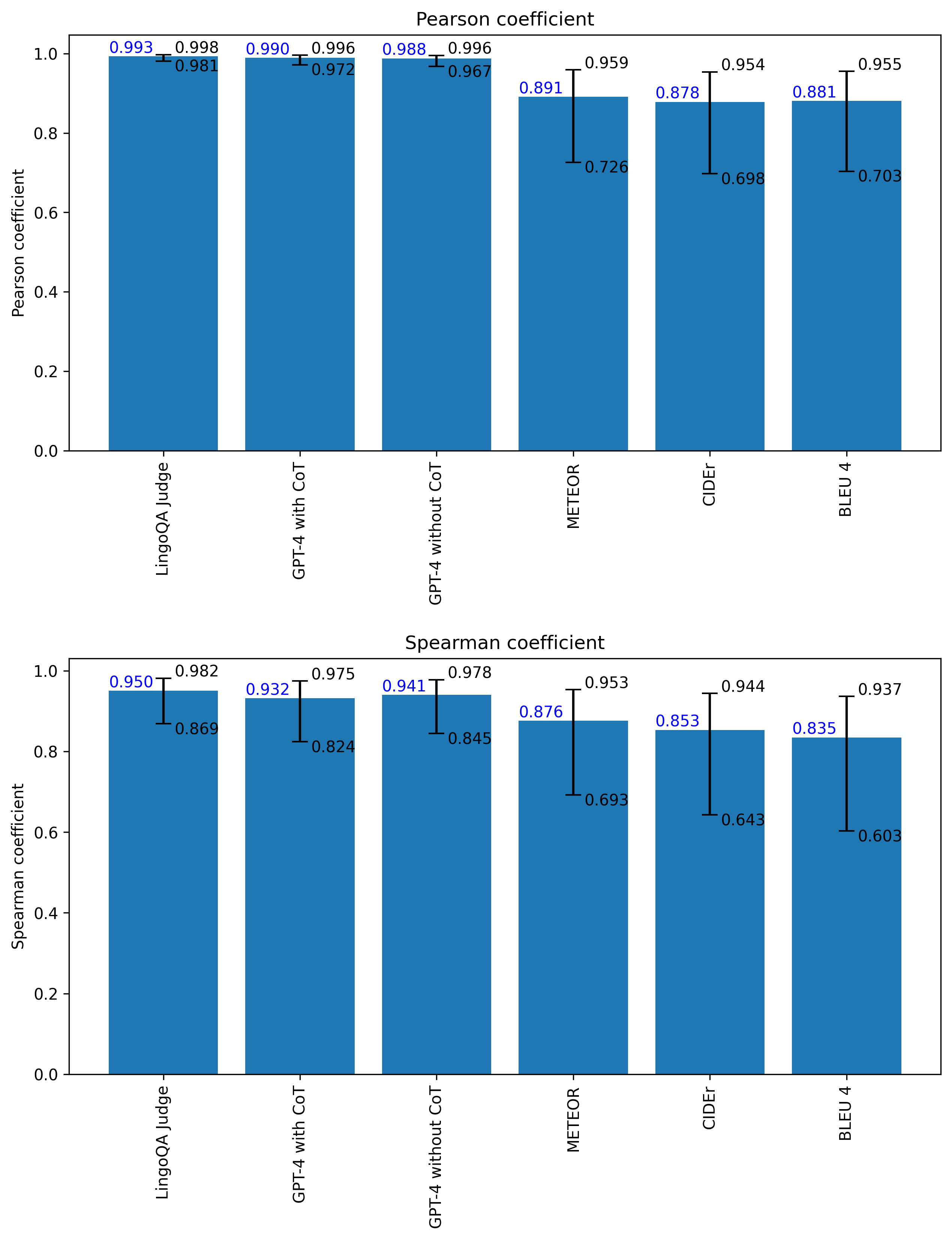}
\caption{\textbf{Correlation coefficients}. Correlation coefficients of the average grade of different models vs. the average human-grades, for different metrics.}
\label{fig:correlation_coefficients}
\end{figure}

In Figure \ref{fig:correlation_trends}, the metric scores are plotted against the human evaluators' grades (from 0 to 1). In red is the least-squares regression of the linear relationship between the metric and the human-assigned grades. Figure \ref{fig:correlation_coefficients} shows the value of both correlation coefficients for each of the 5 metrics, as well as their confidence interval bounds. We note that not only does Lingo-Judge provide higher correlation, it also provides tighter confidence intervals than the other metrics.

\section{Lingo-Judge Generalisation}\label{appendix:generalisation}

To investigate the generalisation abilities of the Lingo-Judge, we examine the performance of the model on a range of answer styles. In particular, we evaluate vision-language models with varying architectures namely GPT4-V, LLaVA and FUYU. We also employ human labelling to obtain ground truth performance of these models. Table \ref{tab:limitations_judge} shows the performance of the Lingo-Judge as measured by the validation accuracy, as well as human, Lingo-Judge and GPT-4 ratingd for comparison. The Lingo-Judge rates FUYU at 45.4\%, compared to 17.69\% as rated by humans. However, the Lingo-Judge rates GPT-4V concise at 56.6\% compared to 56.67\% by humans. This highlights that the model performs the best on short, concise answers, akin to those of humans. 

\begin{table}[tb]
\centering
\caption{\textbf{Robustness to response styles.} Investigation into the impact of the response style on validation accuracy. The Lingo-Judge accuracy is limited on mostly incorrect but long-form answers, with the main failure mode being that these models are rated higher than they should be, compared to the human evaluation.}
\resizebox{0.75\linewidth}{!}{%
\begin{tabular}{@{}lccc|c@{}}
\toprule
& Human & Lingo-Judge & GPT-4 & Lingo-J Val. Acc. \\
\midrule
\textbf{LingoQA} & 57.10 & 60.8 & 3.30 & 89.50 \\
\textbf{GPT-4V} \\
\textit{few-shot} (FS) & 57.69 & 64.6 & 3.39 & 83.27 \\
\textit{concise} (C) & 56.67 & 59.6 & 3.19 & 81.63 \\
\textit{unprompted} (U) & 54.61 & 59.6 & 3.24 & 83.06 \\
\textit{incorrect} (I) & 17.69 & 26.2 & 1.55 & 89.59 \\
\textbf{LLaVA} \\
\textit{concise} (C) & 38.97 & 49.4 & 2.45 & 81.43 \\
\textit{unprompted} (U) & 28.28 & 41.8 & 2.69 & 78.12 \\ 
\textbf{FUYU} & 17.69 & 45.4 & 2.28 & 64.89 \\
\bottomrule
\end{tabular}}
\label{tab:limitations_judge}
\end{table}

\begin{table*}[htb!]
\centering
\caption{\textbf{Correlation study metrics.} Metrics from different models on our evaluation dataset used in the correlation study in Table \ref{tab:correlation}. For reference, we also present metrics for answers provided by human labellers. ``Human'' is the average of inference output scores in range $\left[0, 1\right]$ where 0 is worst and 1 is best, as described in section \ref{sec:evaluation_metric}.}
\resizebox{\linewidth}{!}{
\begin{tabular}{llcccccc}
\toprule
& & Lingo-Judge [\%] $\uparrow$ & BLEU $\uparrow$ & METEOR $\uparrow$ & CIDEr $\uparrow$ & GPT-4 $\uparrow$ & Human $\uparrow$ \\
\midrule
\multirow{13}{*}{\textbf{\textit{Models}}} 
& Model A & $59.6$ & $15.45$ & $18.36$ & $66.32$ & $3.23$ & $0.571$ \\
& Model B & $59.6$ & $15.16$ & $18.84$ & $65.11$ & $3.16$ & $0.564$ \\
& Model C & $57.4$ & $14.87$ & $18.52$ & $65.49$ & $3.08$ & $0.563$ \\
& Model D & $58.2$ & $14.51$ & $18.59$ & $66.02$ & $3.15$ & $0.559$\\
& Model E & $59.0$ & $14.42$ & $18.58$ & $66.95$ & $3.14$ & $0.553$\\
& Model F & $58.0$ & $14.82$ & $18.89$ & $65.43$ & $3.11$ & $0.552$ \\ 
& Model G & $54.8$ & $14.41$ & $17.86$ & $64.67$ & $2.98$ & $0.529$ \\
& Model H & $50.0$ & $13.29$ & $17.44$ & $59.87$ & $2.88$ & $0.520$ \\
& Model I & $53.0$ & $14.63$ & $17.98$ & $64.45$ & $2.96$ & $0.510$ \\
& Model J & $52.6$ & $12.17$ & $17.59$ & $50.45$ & $3.00$ & $0.509$ \\
& Model K & $53.0$ & $13.20$ & $18.03$ & $54.90$ & $3.04$ & $0.500$ \\
& Model L & $51.2$ & $14.69$ & $17.83$ & $64.51$ & $2.91$ & $0.485$\\
& Model M & $43.2$ & $13.76$ & $17.37$ & $60.36$ & $2.67$ & $0.371$\\
& Model N & $35.8$ & $13.18$ & $15.67$ & $56.07$ & $2.41$ & $0.361$ \\
& Model O & $33.6$ & $8.33$ & $14.33$ & $39.16$ & $2.07$ & $0.279$ \\ 
\midrule
\multirow{2}{*}{\textbf{\textit{Humans}}}
& Human labellers group A & $96.6$ & $81.04$ & $52.92$ & $361.77$ & $4.68$ & $0.934$\\
& Human labellers group B & $91.2$ & $61.72$ & $42.57$ & $267.87$ & $4.3$ & $0.894$ \\
\bottomrule
\end{tabular}}
\label{correlation_table_models}
\end{table*}

\section{Training Parameters}\label{training:parameters}
In this sections we present further details on the training parameters used for the LingoQA Baseline. The training process consists of a pre-training stage, and a fine-tuning stage. Table \ref{tab:training_parameters} shows the parameters for pre-training and fine-tuning respectively. The datasets are sampled with equal weight for both pre-training and fine-tuning. The overall training time was ~20h for pre-training and ~5h for fine-tuning on an NVIDIA A100 8GPU 80GB machine.

\begin{table}[htb]
 \centering
 \caption{\textbf{Training parameters.} This table shows the training parameters utilised for the pre-training and for the fine-tuning stages respectively.}
 \begin{tabular}{c|c|c}
 \toprule
 Parameter & Pre-training & Fine-tuning \\
 \midrule
 Precision & bf16 & bf16 \\
 Warm-up steps & 1000 & 1000 \\
 Maximum steps & 100000 & 10000 \\
 Batch size & 6 & 8 \\
 Gradient acc. steps & 1 & 1 \\
 Learning rate & $5 * 10 ^{-5}$ & $5 * 10 ^{-5}$ \\
 Learning rate scheduler & cosine & cosine \\ 
 Weight decay & 0.1 & 0.1 \\
 \bottomrule
 \end{tabular}
 \label{tab:training_parameters}
 \end{table}

\section{LingoQA Baseline Examples}\label{model:examples}

\begin{figure*}[h!]
\centering
\includegraphics[width=0.95\textwidth]{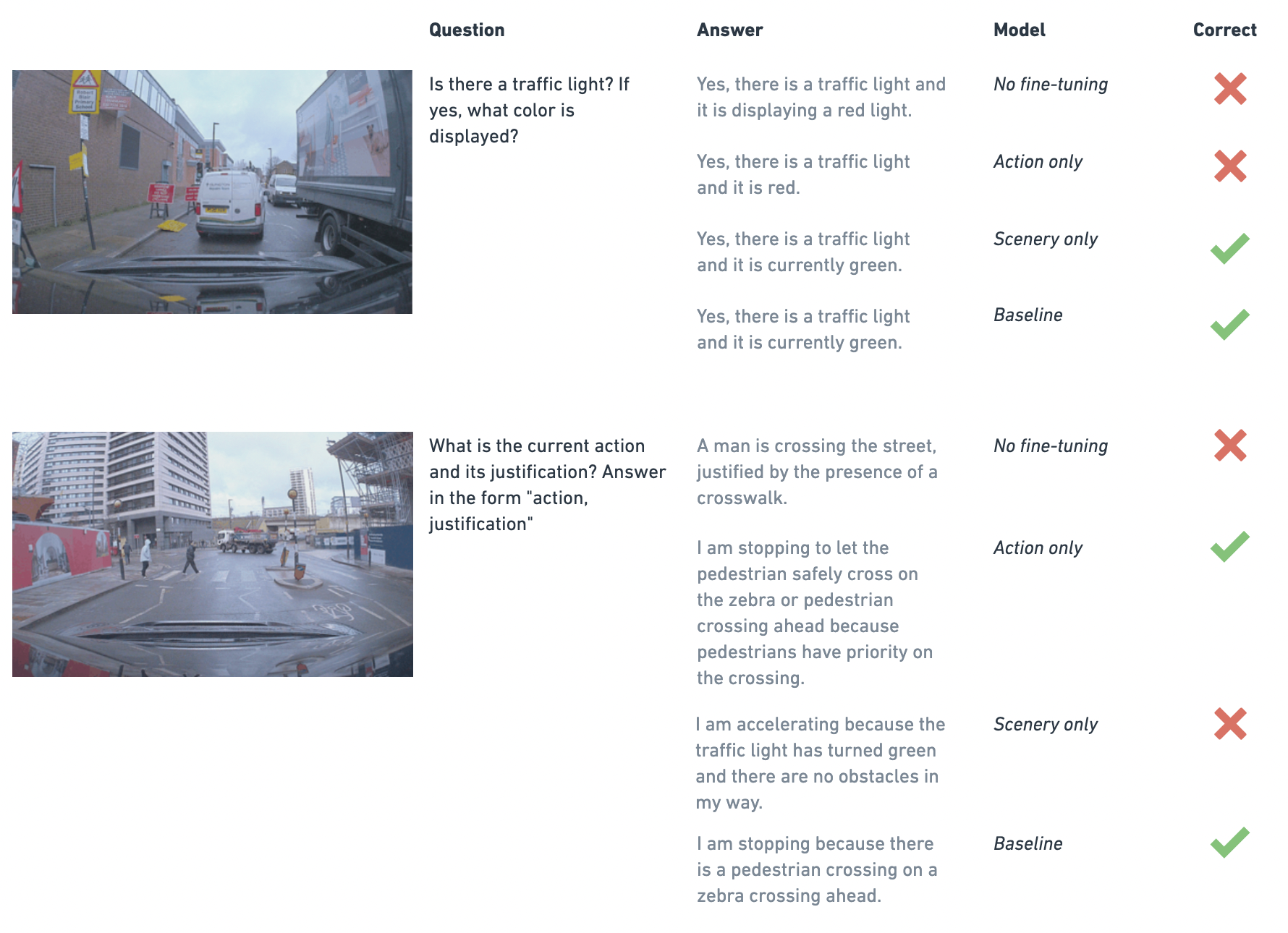}
\caption{\textbf{Examples of model outputs on the LingoQA benchmark.} We compare the baseline with a model that has not been fine-tuned on the LingoQA dataset, a model fine-tuned on the action dataset only, and a model fine-tuned on the scenery dataset only. This shows qualitatively how the baseline can handle both action justification as well as descriptive tasks by combining the strengths of both datasets.}
\label{fig:baseline_examples}
\end{figure*} 

We qualitatively showcase the impact of our proposed LingoQA dataset. Figure \ref{fig:baseline_examples} compares three models: a model that is not fine-tuned on any LingoQA datasets, one that is fine-tuned on the \textit{action} dataset only, one on the \textit{scenery} dataset only, and the baseline that is trained with both. Two questions are asked, one focused on perception only, and one focused on action justification. The action only model performs well at answering action-related questions, but not perception. The scenery only model performs well at perception tasks, but not action justification. The baseline exhibits good performance on both.

\end{document}